\documentclass[authorversion,sigconf]{acmart}
\AtBeginDocument{%
  }

\usepackage{stfloats}
\usepackage{placeins}

\usepackage{multirow}

\usepackage{fancyhdr}
\AtBeginDocument{%
    \addtolength{\footskip}{2.0\baselineskip}%
    \fancyfoot[L]{\textit{\textbf{Preprint --- Author Accepted Version.}}}%
}

\copyrightyear{2025} 
\acmYear{2025} 
\setcopyright{acmlicensed}\acmConference[IVA Adjunct '25]{ACM International
 Conference on Intelligent Virtual Agents}{September 16--19, 2025}{Berlin,
 Germany}
 \acmBooktitle{ACM International Conference on Intelligent Virtual Agents (IVA
 Adjunct '25), September 16--19, 2025, Berlin, Germany}
 \acmDOI{10.1145/3742886.3756718}
 \acmISBN{979-8-4007-1996-7/2025/09}

\begin{document}

\title{The Importance of Facial Features in Vision-based Sign Language Recognition:
Eyes, Mouth or Full Face?}



\author{Dinh Nam Pham}
\orcid{0000-0002-9431-5614}
\affiliation{%
  \institution{German Research Center for Artificial Intelligence (DFKI)}
  \city{Berlin}
  \country{Germany}
}
\email{dinh_nam.pham@dfki.de}

\author{Eleftherios Avramidis}
\orcid{0000-0002-5671-573X}
\affiliation{%
  \institution{German Research Center for Artificial Intelligence (DFKI)}
  \city{Berlin}
  \country{Germany}
}
\email{eleftherios.avramidis@dfki.de}




\begin{abstract}
Non-manual facial features play a crucial role in sign language communication, yet their importance in automatic sign language recognition (ASLR) remains underexplored. While prior studies have shown that incorporating facial features can improve recognition, related work often relies on hand-crafted feature extraction and fails to go beyond the comparison of manual features versus the combination of manual and facial features. In this work, we systematically investigate the contribution of distinct facial regions—eyes, mouth, and full face—using two different deep learning models (a CNN-based model and a transformer-based model) trained on an SLR dataset of isolated signs with randomly selected classes. Through quantitative performance and qualitative saliency map evaluation, we reveal that the mouth is the most important non-manual facial feature, significantly improving accuracy. Our findings highlight the necessity of incorporating facial features in ASLR.
\end{abstract}




\maketitle

\section{Introduction}
\label{sec:intro}

Sign language (SL) serves as a vital means of communication for deaf and hard-of-hearing communities worldwide. However, similar to spoken languages, sign language is not universal. Different countries and regions have their own native sign languages, each with unique grammatical and linguistic structures. Furthermore, there are significant communication barriers between sign language users and those unfamiliar with SL. In an effort to address this, automatic sign language recognition (ASLR) has gained growing attention in recent years, with the number of available datasets and studies increasing. 

As a visual language, sign language conveys information through multiple channels, which can be broadly categorized into (a) manual and (b) non-manual features. Manual features are related to hands and include hand shapes, palm orientation, hand location, and hand movement. In contrast, non-manual features describe expressions which do not involve the hands, such as movements of the mouth, cheeks, head, shoulders, and eyebrows. Sign languages have complex grammars that include non-manual markers, making them essential for understanding SL \cite{nonmanual}. Facial non-manual features are an essential part of SL grammar, not just visual cues. They serve core linguistic functions, with upper-face markers being particularly essential for syntactic and prosodic structures across different sign languages, while the mouth plays a crucial role for the lexical and morphological levels \cite{Pendzich+2020}. For example, eye gaze can mark agreement in relation to agreement verbs, while facial expressions like sucked-in or blown-out cheeks modify signs such as HOUSE to indicate size \cite{nonmanual}. In German Sign Language (DGS), a head nod can mark past tense, and mood is often expressed through facial expressions. Additionally, some signs, like TRAURIG (sad) in DGS, require specific facial expressions to convey their meaning accurately \cite{nonmanual}. Mouth actions can also help disambiguate signs with ambiguous manual features \cite{PhamCA23}. 

Despite this crucial role of non-manual features, only 6\% of sign language recognition (SLR) results from 2015 to 2020 made use of the head, while 5\% utilized mouth features, and 3\% incorporated the eyes \cite{koller_survey}. While recent advances in deep learning have significantly improved manual sign recognition, the extent to which facial features contribute to recognition performance remains underexplored. Given their importance in disambiguating signs and conveying grammatical information, a deeper understanding of their role in vision-based SLR is crucial. Unlike methods that rely on keypoints, skeleton tracking, or handcrafted features extracted from pose estimation models, our approach focuses on end-to-end deep learning-based SLR, where models learn directly from raw video input without performing explicit feature engineering. Additionally, we decided against approaches that depend on external modalities such as depth sensors, wearable gloves, or intrusive tracking devices, making our approach more signer-centric, practical and adaptable to real-world applications.

In this study, we systematically investigate the contribution of different facial regions—eyes, mouth, and full face—to SLR performance. By analyzing classification accuracy and saliency maps on an isolated sign language dataset, we aim to determine which facial features are most informative and relevant for deep learning models in video-based SLR. Furthermore, we explore whether combining facial and manual features improves recognition performance, providing insights into the integration of non-manual signals in ASLR systems. 

\vspace{-1mm}

\section{Related Work}

While there have been works that specifically modeled the facial features as a parameter for SLR, the overwhelming majority rely on full-frame inputs \cite{sarhan_survey, koller_survey}. In such cases, models may implicitly learn to utilize facial cues, but the specific contribution of these features to overall performance remains understudied. For this, studies comparing SLR performance between using solely manual features and combining manual features with facial features could help to understand their significance better.

In \cite{KUMAR201830}, the authors combine hand gestures with facial expressions, which slightly outperforms the model relying only on hands. They extract 3D feature points from the hands and face using Leap Motion and Kinect sensors, then employ a hidden Markov model for each modality to recognize signs. The decisions from both modalities are combined using a Bayesian classifier. This approach achieves relatively small recognition rate gains, but its reliance on sensors limits its practicality for real-world scenarios and makes it less comparable to methods that use only RGB video data. 

As one of the earliest studies to address the significance of facial features in ASLR, \cite{agris} reported improved recognition performance when combining facial and manual features for both isolated and continuous SLR. The authors extracted hand and facial features as coordinates, representing the face as a 16-dimensional vector and the hands as a 22-dimensional vector.

Meanwhile, the authors of \cite{mukushev-etal-2020-evaluation:lrec} conducted a more fine-grained quantitative comparison. They evaluated not only manual features versus the combination of manual and facial features, but also examined specific subsets of facial features: 'Manual \& only eyebrows,' 'Manual \& only mouth,' 'Manual \& eyebrows, mouth,' and 'Manual \& face, eyebrows, mouth.' Using OpenPose \cite{OpenPose}, keypoints were extracted and classified with logistic regression. However, the dataset was explicitly designed to consist of signs that are similar in manual articulation and are distinguished by using non-manual features, leading to non-manual features achieving a mean accuracy of 77\%, compared to 73.4\% for manual features alone. The fact that the dataset solely consisted of cases where non-manual features play a decisive role by design, diminishes the effect of highlighting their importance in general.

Similarly, \cite{PhamCA23} addressed the function of the mouth to disambiguate such homonyms, based on a dataset of pairs with identical or similar manual signs, but different meanings. The authors demonstrated that the combination of mouth and manual features outperform the model using only manual features. Unlike the other works described before, this approach avoided hand-crafted feature extraction, using a deep learning model instead to classify the video streams end-to-end. Each video stream was processed by a separate instance of the same model, with hand and mouth features combined in a late fusion fashion.

Although not primarily aimed to address the importance of facial features in SLR, a deep learning model was trained in \cite{Faceswap} with full-frame inputs. The results interestingly demonstrated that omitting the face in the full-frame significantly decreased the accuracy whereas the utilization of face swapping as a data augmentation method lead to accuracy gains, further underscoring the impact of facial features in SLR.

In essence, previous studies have primarily evaluated the impact of combining manual and facial features. They often relied on hand-crafted features, though deep learning approaches were also used. While all works reported accuracy gains from incorporating facial features, the extent of improvement varied, particularly in datasets designed to emphasize non-manual features. Contrary to that, in this work, we construct an isolated SLR dataset with randomly selected classes, independent of their co-occurrence with non-manual articulation. We focus on deep learning models, which better reflect the currently commonly used methods \cite{sarhan_survey}, and conduct a detailed investigation into the contribution of the eyes, mouth, and full face. By quantitatively evaluating their integration with manual features and qualitatively analyzing saliency maps, we offer new insights into the importance of facial features in ASLR.

As this study does not include direct comparisons to other sign language recognition methods, our objective is not to achieve state-of-the-art accuracy, but to provide a focused and controlled analysis of the contribution of facial regions using modern vision models. Most prior works that aim to highlight the significance of non-manual features combine manual and facial features without isolating specific facial regions, or rely on hand-crafted features and sensor data—methods that are increasingly obsolete given the current dominance of end-to-end deep learning SLR models for RGB video analysis. In contrast, our approach uses state-of-the-art deep learning backbones under a consistent training setup, enabling a reproducible and architecture-agnostic evaluation of non-manual feature importance. We believe that these controlled conditions offer a complementary perspective, and our findings can inform the integration of facial cues into future multimodal SLR systems.

\section{Dataset}

In order to assess the importance of facial features in ASLR, a dataset is required to train and evaluate models. 
For this purpose, the Public DGS Corpus \cite{dgscorpus_3} was selected to create a dataset for isolated sign language recognition with glosses as class labels. The Public DGS Corpus is well-suited for this task due to its comprehensive annotations, diversity of speakers as well as the quality and quantity of video recordings. Data, including 550 hours of DGS signing, were collected from 330 signers across 12 different locations in Germany, ensuring a balanced representation of age, gender, and regional variation among participants \cite{hanke:20016:sign-lang:lrec}. Using the gloss annotations, we randomly selected 12 glosses, each occurring at least 500 times, to ensure a sufficient number of instances for robust model training. Using the timestamps associated with these annotations, we extracted the corresponding video clips from the recordings. Each instance in the dataset consists of a video clip displaying a signer performing the sign for a single gloss. In our dataset, the glosses serve as the class labels, while the individual video clips represent the instances within each class. 

\begin{figure}[] 
    \centering
    \includegraphics[width=0.48\textwidth,clip,trim=0 0 0 25]{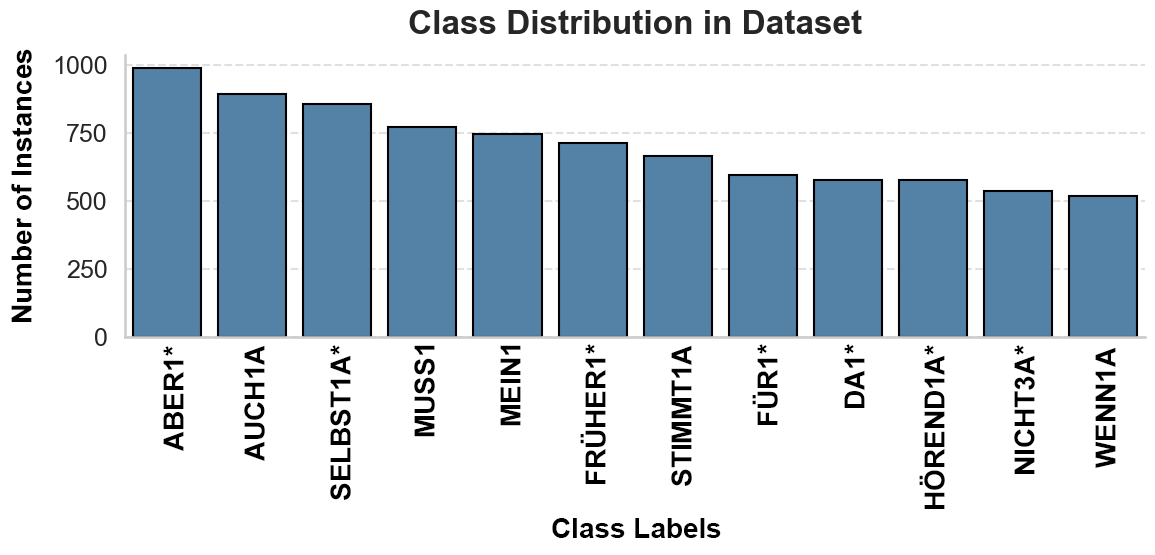}
    \caption{Class distribution in the dataset.}
    \label{fig:Class_distribution}
\end{figure}

The class distribution is illustrated in Figure \ref{fig:Class_distribution}. The class with the highest amount of instances contains 989 instances, while the gloss with the fewest instances includes 518 videos. Given that this imbalance ratio is less than 1:2 and considering that most studies on class imbalance focus on ratios starting from 1:4 \cite{Krawczyk2016}, we consider the dataset to not be severely imbalanced. As a result, no additional measures were taken to adjust the class distribution. 

In accordance with the annotation standard of the Public DGS corpus ~\cite{AP_AnnotationConventions}, the names of the labels begin with a gloss, i.e. a word of the spoken language that represents the core meaning of a specific sign. This is followed by a number that describes a lexical variant and a letter denoting a phonological variant. An optional asterisk indicates that the observed form of the sign differs slightly from its standard or citation form.

We split the dataset into training, validation and test sets using an 8:1:1 ratio while preserving the class distribution in each set. To address variations in video frame length, we normalized the videos by padding them with repetitions of the last frame until each video reached a uniform length of 32 frames. To evaluate the importance of different facial regions, we constructed four versions of the dataset by cropping each video to specific regions of interest (ROIs): the eyes, mouth, full face, and body. For cropping the mouth region, we adopted the implementation from \cite{Ma2022}. The face region was extracted using the Face Recognition library\footnote{Adam Geitgey, \textit{Face Recognition}, GitHub. [Online] Available: \url{https://github.com/ageitgey/face_recognition}.}, while MediaPipe~\cite{mediapipe} was employed to detect the coordinates of the eyes, upper body, and hands for cropping. The upper body ROI was defined by determining the outermost coordinates from the detected upper body and hand landmarks, ensuring that both the body and hands were always fully visible. Finally, the videos of the ROIs were resized to 224x224 pixels since this is a common image size for pre-trained vision models.

\begin{figure}[] 
    \centering
    \includegraphics[width=0.47\textwidth]{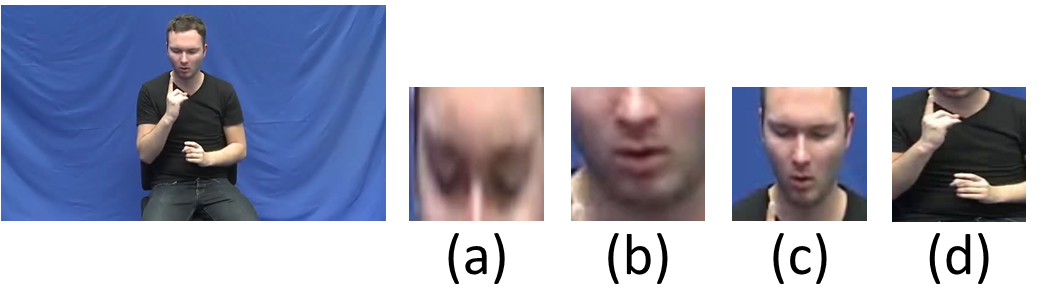}
    \caption{An example of the dataset \cite{dgscorpus_3} being cropped to the regions of interest: (a) eyes, (b) mouth, (c) face, and (d) body.}
    \label{fig:img1}
\end{figure}

\section{Experiments}

For the ASLR, we leveraged two distinct state-of-the-art architectures: Channel-Separated Convolutional Networks (CSN) \cite{csn} and Multiscale Vision Transformers (MViT) \cite{mvit}. More specifically, we used the CSN R101 and MViT 32x3 model implementations from PyTorchVideo~\cite{Pytorchvideo}, which were pretrained on the Kinetics-400 \cite{kay2017kineticshumanactionvideo} dataset under the same frame length and image size setting as the items in our dataset. The CSN architecture factorizes 3D convolutions, whereas MViT integrates a multiscale feature hierarchy with the transformer architecture. 

The reason we chose these models is that they differ fundamentally from each other in their design principles (convolutional vs. transformer-based), allowing us to assess whether observed trends generalize across architectures and gain complementary insights into the significance of different regions of interest. By comparing the performance and feature utilization of both architectures, we can identify consistent patterns in how facial features contribute to sign language recognition, leading to conclusions about the role of facial non-manual signals in vision-based models that can be generalized across the current state of the art. The use of the two models for SLR was also supported by the fact that they have demonstrated state-of-the-art performance on human action recognition, as they have outperformed successful and commonly used SLR architectures, such as I3D and R(2+1)D \cite{gao2022,han2022,Li2019,sarhan2020,fink2021,Faceswap}, on the Kinetics400 benchmark \cite{kay2017kineticshumanactionvideo}. 

\begin{figure}[] 
    \centering
    \includegraphics[width=0.4\textwidth]{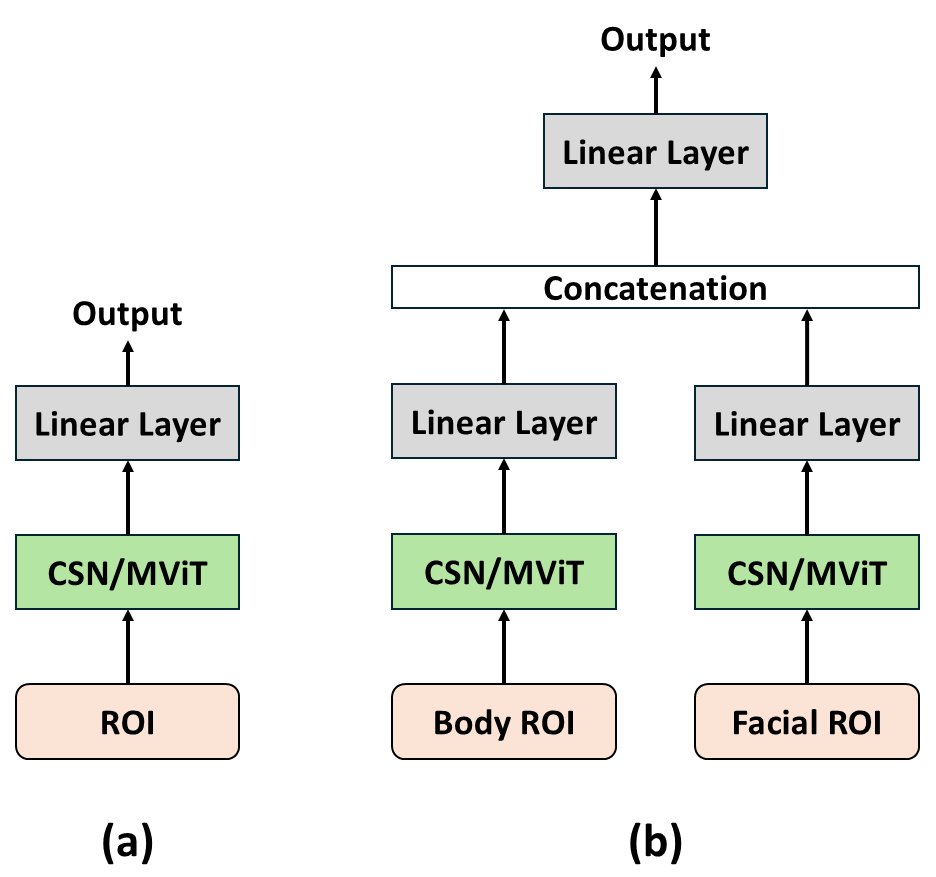}
    \caption{Model architectures for (a) single ROI and (b) two ROIs (in this example \emph{body + facial}).}
    \label{fig:Models}
\end{figure}

We trained both CSN and MViT on each of the four ROIs. For both architectures, the output layer was replaced with a linear layer with an output size of 12, corresponding to the number of class labels. To evaluate the impact of facial features on SLR with manual signs, we combined the body ROI with each facial ROI (eyes, mouth, and face). For each model $M \in \{CSN, MViT\}$ and each facial ROI $R_F \in \{Eyes, Mouth, Face\}$, we constructed a new model where one instance of $M$ processes $R_F$ and another processes the body ROI. The outputs of both instances were fused and propagated to a linear output layer. This late-fusion approach integrates facial non-manual features with manual signals, enabling observation to what extent they improve the performance. This architecture is illustrated in Figure \ref{fig:Models}.

For all experiments, the same hyperparameters and settings were employed. We used the cross-entropy loss, the Adam optimizer, a learning rate of $10^{-5}$, and a batch size of 3 due to limited memory capacity. Additionally, RandAugment \cite{randaug} with a magnitude of 4 and $N=3$ was applied as a data augmentation technique to each video in the training set before it was fed into a model. The models were trained on the training set and validated on the validation set for 100 epochs. The weights yielding the highest top-1 accuracy on the validation set were used for final evaluation on the test set.

\begin{table*}[t]
\centering
\begin{tabular}{l|ccc|ccc}
\toprule
\multirow{2}{*}{Region of Interest} & \multicolumn{3}{c|}{CSN} & \multicolumn{3}{c}{MViT} \\  
                                    & Top-1 Acc.  & Top-3 Acc.  & F1-Score  & Top-1 Acc.  & Top-3 Acc.  & F1-Score  \\  
\midrule
Eyes                                & $20.59 \pm 2.97$   & $44.82 \pm 3.65$     & 0.1853     & $22.41 \pm 3.06$     & $46.36 \pm 3.66$     & 0.2006     \\  
Mouth                               & $48.18 \pm 3.67$    & $76.19 \pm 3.12$     & 0.4864     & $51.40 \pm 3.67$     & $77.31 \pm 3.07$    & 0.5067     \\  
Face                                & $60.64 \pm 3.58$    & $82.07 \pm 2.81$     & 0.6148     & $59.66 \pm 3.60$     & $79.55 \pm 2.96$     & 0.6002     \\  
Body                                & $80.53 \pm 2.90$    & $94.96 \pm 1.60$     & 0.8164     & $84.03 \pm 2.69$    & $93.42 \pm 1.82$     & 0.8496     \\  
\midrule
Body + Eyes                         & $82.77 \pm 2.77$     & $95.24 \pm 1.56$     & 0.8400     & $82.91 \pm 2.76$     & $94.54 \pm 1.67$     & 0.8485     \\  
Body + Mouth                        & $\textbf{88.24} \pm 2.36$    & $97.20 \pm 1.21$     & \textbf{0.8890}     & $86.42 \pm 2.51$     & $\textbf{97.76} \pm 1.09$     & 0.8735     \\  
Body + Face                         & $87.26 \pm 2.45$     & $97.62 \pm 1.12$     & 0.8790     & $86.98 \pm 2.47$     & $96.22 \pm 1.40$     & 0.8767     \\  
\bottomrule
\end{tabular}
\caption{Top-1 accuracy, top-3 accuracy and F1-Score for CSN and MViT across different regions of interest.}
\label{tab:roi_results}
\end{table*}

\section{Results and Discussion}

\subsection{Classification Performance}

We report the experimental results in Table \ref{tab:roi_results}, including accuracies with a confidence interval of 95\% and F1-scores for both CSN and MViT. It is striking that CSN and MViT perform relatively similarly across all metrics and ROIs. Unsurprisingly, the eyes on their own achieved the lowest accuracies. However, with top-1 accuracies of 20.59\% and 22.41\%, it seems to contain useful information to some extent. The full face ROI strongly outperforms the other facial ROIs, which makes sense, since it contains both these facial ROIs and other features such as head pose and facial expressions as well. The body ROI, demonstrating the best performance across all ROIs, when using only one video stream, was also expected.

The accuracies obtained by combining eyes and body are statistically not significantly different from the models using only the body as the confidence intervals overlap. This suggests that while the eyes can provide useful features, their contribution becomes relatively insignificant when manual features are present—at least in this experimental setting. In contrast to that, the incorporation of the mouth as well as the face significantly improves the top-1 accuracy of CSN with the body ROI. However, for the MViT model, as its top-1 accuracy for the body ROI was already relatively high, a statistically significant difference cannot be observed for the "body + mouth" and "body + face" ROIs. Only MViT's top-3 accuracy was significantly improved by adding the mouth. 

Interestingly, while the face outperformed the mouth ROI when using them on their own, the performance difference between them, when combined with the body, becomes insignificant. With overlapping confidence intervals across all accuracies and models, the results do not indicate a clear distinction between the two ROIs when the manual signals are present. This may indicate that \textit{the mouth area is the most important facial feature}. 

Furthermore, even though the addition of mouth and face significantly improves the top-1 accuracy of CSN compared to solely using the body, they fail to make a significant difference for the top-3 accuracy. This suggests that \textit{the facial features can make a difference for vision-based models to better distinguish between top 3 candidates}. This aligns with the known functionality of non-manual signals, such as mouthing, to disambiguate between signs that share the same manual articulation but differ in meaning \cite{PhamCA23}.

In an effort to explain why the inclusion of the mouth ROI increases the performance when modeled together with the body ROI, we use some linguistic observations. The signs for 6 of our classes (ABER, AUCH, SELBST, MUSS, NICHT, WENN) are based on the same handform (extended index finger) but with different orientation, position or movement. An assumption is that from a computer vision perspective, the respective video clips are similar (e.g. share identical frames when taken from different angles), so that the mouthings, which depict different vowels, help the models make the distinction.
\FloatBarrier
\subsection{Saliency Maps}

\begin{figure*} 
    \centering
    \includegraphics[width=0.683\textwidth]{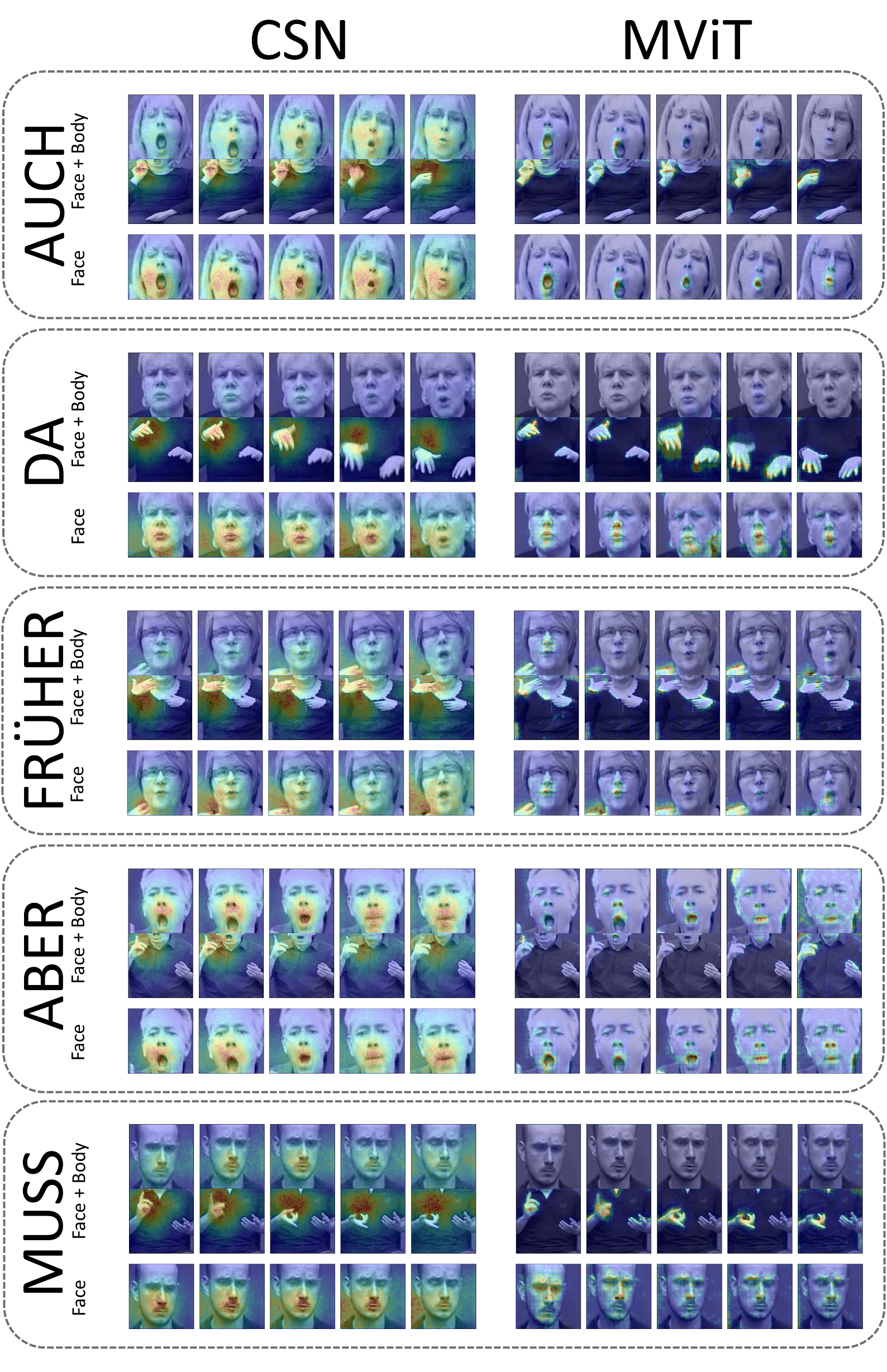}
    \caption{Saliency maps for some videos in the test set \cite{dgscorpus_3}, visualising the models for the face as well as the "body + face" fusion.}
    \label{fig:saliency}
\end{figure*}

In addition to the quantitative results, we generated saliency maps to visualize the most relevant regions for both models, providing further insights into the importance of facial features in a qualitative manner. For this, we computed vanilla gradient saliency maps \cite{saliency} for videos from each class and applied SmoothGrad \cite{smoothgrad} with a sample size of 50 to reduce noise in the pixel attributions. Our analysis focused on two regions of interest: (1) the face alone and (2) the combination of the face and body. This approach allowed us to identify the most critical facial regions and examine how their relevance changes when contextual information from the body is included, which is of particular interest in this study. Examples are shown in Figure \ref{fig:saliency}, which illustrates the first, fourth, seventh, tenth, and 13th frames of a single instance for a few classes. The top 0.5\% attribution values were set to 1 while the remaining values were normalized relative to these outliers. 

Throughout all classes, the saliency maps of MViT appear to be much more fine-grained compared to those of CSN. This likely stems from the underlying architecture differences between the two models. MViT, with its hierarchical attention mechanism, is better equipped to capture localized and detailed spatial-temporal dependencies, resulting in finer feature attribution. Despite these different granularities, both models highlight the same crucial regions, making the observations more meaningful. For the face ROI, the pixels of the mouth area were consistently highlighted as the most relevant features, reinforcing the quantitative findings that the mouth is the most important facial feature. This aligns with the findings of eye-tracking studies where signers were observed to not only mainly focus on the face during sign language comprehension, but also primarily fixate on the mouth area while doing so~\cite{Mastrantuono2017,Emmorey2009-fe}. 
However, some distinctions exist in instances. For example, CSN sometimes highlights broader facial regions, including the cheeks, nose, and jawline, whereas MViT concentrates on finer details, particularly the lips. Moreover, in a noteworthy number of instances, the nose and eyes were significantly marked, speaking for the relevance of these features. 

Furthermore, we often observed that the addition of the body leads to a diminished saliency of the face. This effect is particularly evident in the example of the MViT model for the class "DA" as shown in Figure \ref{fig:saliency} where the face region is nearly absent from the saliency map, shifting attention almost entirely to the hands. When it comes to the body ROI, we noticed that the saliency of the right hand was frequently higher than for the left hand, which aligns with the work of Fink et al. \cite{fink2021} who made the same observation on saliency maps for the I3D model on a French Belgian Sign Language dataset. We share their interpretation that the right hand is of bigger importance for models as most signers use their dominant hand to sign. This bias could lead to lower accuracy for left-handed signers and should be considered in future works by making SL dataset efforts or including horizontal flipping in data augmentation to mitigate this. Furthermore, in cases where the hands appear within the face ROI, such as in the "FRÜHER" class in Figure \ref{fig:saliency}, the models strongly focused on the hands rather than the face. This suggests that even when trained primarily on facial features, the models implicitly recognize the relevance of manual features when they appear in the frame. 

While our study mainly focused on facial features, it is important to acknowledge that non-facial non-manual features (e.g., head pose, shoulder movement, and torso orientation) may also play a significant role in ASLR. As our body ROI includes the upper body and hands, it primarily captures manual articulators (i.e., the hands and their movements) along with their spatial positioning relative to the signer’s body. As such, signals like subtle shoulder shifts or head tilts are not explicitly disentangled in our analysis. Future work should explore these non-facial non-manual cues more explicitly to better understand their contribution and potential interaction with facial and manual features.

\section{Conclusion}

In this paper, we systematically investigated the role of facial features in vision-based sign language recognition using CSN and MViT models. Through experiments on an isolated SLR dataset, we demonstrated that incorporating facial features can significantly improve SLR models, aligning with prior works and underscoring the need for future SLR models to integrate non-manual signals. Our approach focuses on state-of-the-art end-to-end deep learning methods as opposed to previous work using manual feature extraction such as keypoints. Our results suggest that the mouth is the most important facial feature when combined with manual features, as there were no statistically significant performance differences when manual features were combined with either the mouth or the full face, across all models and metrics. Additionally, it is shown that the facial features can aid the vision-based models to better distinguish between top candidates. The saliency map analysis further confirmed that the mouth is the most important facial feature, consistently highlighting the mouth as the most relevant facial region. Compared to that, the eyes contributed less significantly, although they appeared to provide some useful cues to some extent. Our findings emphasize the need of incorporating facial non-manual markers—especially mouth actions—into SLR models. 

Future research should explore finer-grained non-manual features such as eye gaze, blinks, nose movement, head pose, tongue position, and cheek articulation, with further guidance from linguistic theory observations. The experiments could be expanded to include more models and perform training and evaluation on a number of different SLR datasets, including a larger number of glosses and continuous SLR, to provide even more robust and generalizable findings. Additionally, leveraging transfer learning from related tasks, such as automatic lip reading \cite{PhamVSRMouth2025} and eye tracking, could advance the integration of non-manual features in SLR systems.

\section*{Ethical Considerations}
In our work, we present experiments on the German Sign Language (DGS), which should be seen and respected as the primary languages of the respective language community. 
While the overarching objective of this research is to promote equitable access to language technologies for sign language users, the predominance of hearing researchers in NLP entails the risk of developments that are not in accordance with the will of the respective communities, and therefore it is required that every research step takes them in constant consideration.
To address this concern, we have included members of the Deaf/deaf and hard-of-hearing communities in our broader research line, as part of the research team, consultants and participants in user studies and workshops, and we have been in co-operation with related unions and communication centers.
It should also be noted that our experiments are part of a broader series of research projects, and the results presented here should be by no means considered ready for production nor used as final products without the agreement of the communities. 
The use of datasets follows their respective licenses and restrictions, and every follow-up work should adhere to those.  

\begin{acks}
The research reported in this paper was supported by BMBF (German Federal Ministry of Education and Research) via the project SocialWear (grant no.~01IW20002).
\end{acks}

\bibliographystyle{ACM-Reference-Format}
\bibliography{main.bib}

\end{document}